\documentclass[conference]{IEEEtran}
\IEEEoverridecommandlockouts
\usepackage{cite}
\usepackage{amsmath,amssymb,amsfonts}
\usepackage{algorithmic}
\usepackage{graphicx}
\usepackage{textcomp}
\usepackage{xcolor}
\def\BibTeX{{\rm B\kern-.05em{\sc i\kern-.025em b}\kern-.08em
    T\kern-.1667em\lower.7ex\hbox{E}\kern-.125emX}}
\makeatletter
\def\ps@IEEEtitlepagestyle{%
  \def\@oddfoot{\mycopyrightnotice}%
  \def\@evenfoot{}%
}
\def\mycopyrightnotice{%
  {\hfill \footnotesize 978-1-7281-6215-7/20/\$31.00 \copyright 2020 IEEE\hfill}
}
\makeatother
\begin{document}

\title{SE-ECGNet: A Multi-scale Deep Residual Network with Squeeze-and-Excitation Module for ECG Signal Classification
}

\author{Haozhen Zhang$^{1}$
\\
\IEEEauthorblockA{\textit{Tianjin University
} \\
\textit{College of Intelligence and Computing
}\\
Tianjin, China
\\
viktor@tju.edu.cn}
\and
$\thanks{$^{1}$ Equal contribution}$
\IEEEauthorblockN{Wei Zhao$^{1}$}
\IEEEauthorblockA{Tianjin University
\\
\textit{College of Intelligence and Computing}\\
Tianjin, China \\
amadeuszhao@tju.edu.cn}
\and
$\thanks{$^{*}$ Corresponding author}$
\IEEEauthorblockN{Shuang Liu$^{*}$}
\IEEEauthorblockA{Tianjin University
\\
\textit{College of Intelligence and Computing}\\
Tianjin, China \\
shuang.liu@edu.tju.cn}
}

\maketitle

\begin{abstract}
The classification of electrocardiogram (ECG) signals, which takes much time and suffers
from a high rate of misjudgment, is recognized as an extremely challenging task for cardiologists. The
major difficulty of the ECG signals classification is caused by the long-term sequence dependencies. Most
existing approaches for ECG signal classification use Recurrent Neural Network models, e.g., LSTM and
GRU, which are unable to extract accurate features for such long sequences. Other approaches utilize
1-Dimensional Convolutional Neural Network (CNN), such as ResNet or its variant, and they can not make
good use of the multi-lead information from ECG signals.Based on the above observations, we develop a multi-scale deep residual network for the ECG
signal classification task. We are the first to propose to treat the multi-lead signal as a 2-dimensional matrix
and combines multi-scale 2-D convolution blocks with 1-D convolution blocks for feature extraction. Our
proposed model achieves 99.2\% F1-score in the MIT-BIH dataset and 89.4\% F1-score in Alibaba dataset
and outperforms the state-of-the-art performance by 2\% and 3\%, respectively,
view related code and data at https://github.com/Amadeuszhao/SE-ECGNet
\end{abstract}

\bibliographystyle{plain}

\section{INTRODUCTION}
It is a fact that cardiovascular diseases (CVDs) are the number one cause of death globally, taking an estimated 17.9 million human lives each year nowadays, 
among which 85\% are due to heart attack and stroke. Therefore detecting and treating heart diseases are of great importance and are getting more and more attention worldwide.

An electrocardiogram (ECG), is a test that measures the electrical activity of the heartbeat. 
The ECG signal provides two major kinds of information, i.e.,  the electrical activity status (normal, slow, fast, irregular) by observing time intervals, and the size of the heart  
by measuring the amount of electrical activity passing through the heart muscle. 
Therefore, a cardiologist can observe the abnormal behaviours of hearts based on ECG signals. However, detecting abnormal ECG signals can be very challenging and time-consuming work for a cardiologist. 
Therefore, active researches~\cite{nasiri2009ecg} have been conducted on the automatic ECG signal classification problem.

Machine learning approaches, such as support vector machine (SVM)~\cite{nasiri2009ecg}, have been employed for ECG signal classification~\cite{nasiri2009ecg}. 
The accuracy of ECG signal classification using machine learning algorithms is much lower than that by a cardiologist due to the high complexity of classification tasks and the poor feature extraction capability of traditional machine-learning algorithms. 

%

With the rapid development of deep learning techniques, more advanced deep learning models, such as ResNet 1-D~\cite{brito2019electrocardiogram}, ECGNet~\cite{murugesan2018ecgnet}, and Seq2Seq~\cite{mousavi2019inter} are employed in the ECG signal classification task. One of the most popular techniques adopted is the time sequence model, e.g., long short-term memory (LSTM)~\cite{yildirim2018novel}, gated recurrent unit (GRU)~\cite{lynn2019deep} and some other variants bases on recurrent neural network (RNN), such as Sequence-to-sequence (seq2seq) models. Due to the long time sequences of ECG signals (a single sample contain 40,000 points), using RNN-based models will lose a large amount of information during the feature extraction process and result in low classification accuracy.  ResNet 1-D~\cite{brito2019electrocardiogram} and ECGNet~\cite{murugesan2018ecgnet}, both use general deep residual networks and 1D convolution kernel for feature extraction. These approaches achieve better results, yet they fail to utilize the multi-head signal in ECG signals. 


ECG signals consist of multiple leads and each lead has an extremely long sequence which is further composed of many sub heartbeat sequences. 
The extremely long signal sequence makes it difficult for traditional feature extraction models such as long-short term memory (LSTM) to capture the feature of the entire sequence. Moreover, current pre-processing techniques~\cite{brito2019electrocardiogram} directly truncate the long sequence into several fix-length subsequences, which may lose the original signal information. 
Existing approaches~\cite{murugesan2018ecgnet} all fail to utilize the multi-head information available in ECG signals, which is potentially able to strengthen useful features in the single-head signal since the multi-head signals share similar trends. 

Based on the above observations, in this work, we propose a multi-scale deep residual neural network called SE-ECGNet, which makes use of the multi-lead ECG signal information and outperforms other state-of-the-art models by a large margin.  Our main contributions are summarized as follows: 
\begin{enumerate}
    \item We propose a new deep learning model structure which (1) utilizes multi-scale, diverse-size kernel convolution blocks in parallel for effective feature extraction; (2) adopts the squeeze-and-excitation module (SE)~\cite{SENet} to focus on more important features; and (3) combines 2-D convolution blocks with 1-D convolution blocks for more effective feature learning. 
    \item We evaluate our model with two datasets, e.g., the MIT-BIH dataset~\cite{moody2001impact} and the Alibaba dataset, and compare our approach with three state-of-the-art models. Our proposed model achieves 99.2\%   F1-score on the MIT-BIH dataset and 89.8\% on the Alibaba dataset, and outperforms the other models by a margin of 2\% and 3\%, respectively. 
\end{enumerate}



The rest of this paper is organized as follows. The related work is discussed in section~\ref{sec:related}. We introduce our model and the evaluation results in section~\ref{sec:model} and section~\ref{sec:exp}, respectively. 
Section~\ref{sec:conclu} concludes the work.
\section{Related work}
\label{sec:related}




The classification of ECG signals is known to be a critical yet challenging task. There have been a large number of approaches~\cite{sharma2016efficient} proposed to solve the task. We classify the existing approaches into two categories, i.e., traditional machine learning based approaches and deep learning based approaches, based on the techniques used for feature learning.   



\subsection{Traditional Machine Learning based Classification Methods}

Approaches adopt traditional machine learning techniques to classify ECG signals usually first extract manually crafted features from ECG signals and then use different classifiers or clustering algorithms to conduct classifications. 
\cite{nasiri2009ecg}  first extract 22 features semi-automatically from time-voltage of R S T P Q features of the ECG signal, and then the Support Vector Machine (SVM)~\cite{cauwenberghs2001incremental} is adopted for classification. The genetic algorithm is used to improve the generalization ability of the SVM classifier. This approach achieves an accuracy of 83.4\%. 
\cite{llamedo2012automatic} propose a patient adaptive ECG signal classification method, which adopts the linear discriminant classifier~\cite{sharma2016efficient} and the Expectation Maximization clustering algorithm~\cite{cimen2016arrhythmia} for effective feature classification. 
%
\cite{gutierrez2017dsp} first exact features using wavelet transform and then classify using the probabilistic neural network, which only includes the fully connected layers, and achieves an accuracy of 92.3\%.  
Traditional machine learning based approaches are restricted by manually crafted features and do not achieve good performance.

\subsection{Deep Learning based Classification Methods} 
Due to the time sequence nature of the ECG signals, it's intuitive to consider using Recurrent Neural Networks (RNN) to learn features. Mousavi and Afghah~\cite{mousavi2019inter} propose to combine Convolutional Neural Network (CNN) and the sequence to sequence (Seq2Seq) model, which has shown good performance on machine translation tasks, to solve the ECG signal classification task. This work first feeds the fixed length and continuous ECG signal into a CNN model. Then a sequence to sequence model is adopted to further learn the feature. The trained encoder is adopted for the classification task.   
%
The approach adopts the public MIT-BIH dataset~\cite{moody2001impact}, and standard pre-processing steps, such as normalization, t wave, R-peak identification, and signal truncation, are conducted~\cite{wavelet1995}. The approach combines several original categories into one new category group (such as taking original category V and E into a new category group named "V"), which reduces the difficulty of the classification task. It achieves an F1-score of 99.5\% on the simplified evaluation setting.

Li et al.~\cite{li2017classification} utilize a 1-Dimensional CNN model to conduct the classification of 5 typical kinds of ECG signals, i.e., normal, left bundle branch block, right bundle branch block, atrial premature contraction, and ventricular premature contraction, on the MIT-BIH dataset. The proposed model consists of five layers, including two convolution layers, two downsampling layers, and one full connection layer. The proposed CNN model achieves a promising classification accuracy of $97.5\%$, which significantly outperforming several traditional machine learning models~\cite{li2017classification}. 

Ng et al.~\cite{Andrew2017} achieve state-of-the-art performance in ECG signal classification using deep Neural Network.   
They use a 34-layers deep residual network, which consists of 16 residual blocks with 2 convolutional layers per block and a max-pooling operation as sub-sampling per two blocks. The network takes a time-series of raw ECG signals as input~\cite{ZioPatchMonitor2017dataset} and outputs a sequence of label predictions. Using F1-score as an evaluation metric, the network reaches a score of 0.809, exceeding the score of 0.751 by a cardiologist. 

Since CNN models are known to be effective on the spatial feature extraction, and RNN models are known to be effective on the time series feature extraction, 
researchers propose to combine CNN and RNN modules for feature extraction~\cite{murugesan2018ecgnet}, \cite{xie2018bidirectional}. 
\cite{murugesan2018ecgnet} uses a wavelet-based pre-processing method~\cite{rajoub2002efficient} to ﬁlter the noise and split the ECG signal to fixed-length based on R-peaks. And then put 1-D sequence data into classic CNN model, such as ResNet~\cite{ResNet}. Then an RNN model, such as LSTM and BiLSTM, is adopted to further learn features from the output of CNN models.
\cite{xie2018bidirectional} propose another way to use features based on Time-Sequence, in which they use LSTM to collect features and then concatenate them with features that CNN extracted. Then a fully connected layer is adopted to classify the results. This model can achieve an accuracy of 97.6\%.



Motivated by the recent advances of the 2-Dimensional CNN model in computer vision, Xie et al.~\cite{salem2018ecg} propose a transfer learning approach, which transfers knowledge (i.e., model parameters) learned on the image classification task to solve the ECG signal classification task. 
%
The approach first transfers the 1-dimensional ECG signals into the spectrogram data and then use the pre-trained DenseNet model~\cite{salem2018ecg} to extract feature maps from the spectrogram data. The extracted feature maps are feed into an SVM model to classify different arrhythmia disease. The approach demonstrates that feature maps learned in the deep neural network, which is trained on great amounts of generic input images, are able to capture the features of the ECG signals.  
This model achieves an accuracy of 97.23\% on the MIT-BIH dataset.  

Our approach first proposes to utilize the multi-lead information of the ECG signal, we combine the 2-D convolution blocks and 1-D convolution blocks, with the multi-scale parallel structure for feature extraction. Compared to state-of-the-art approaches, our approach achieves the best results on both the MIT-BIH dataset and the Alibaba dataset.

\begin{figure}[t]
    \flushleft 
    \includegraphics[scale=0.6]{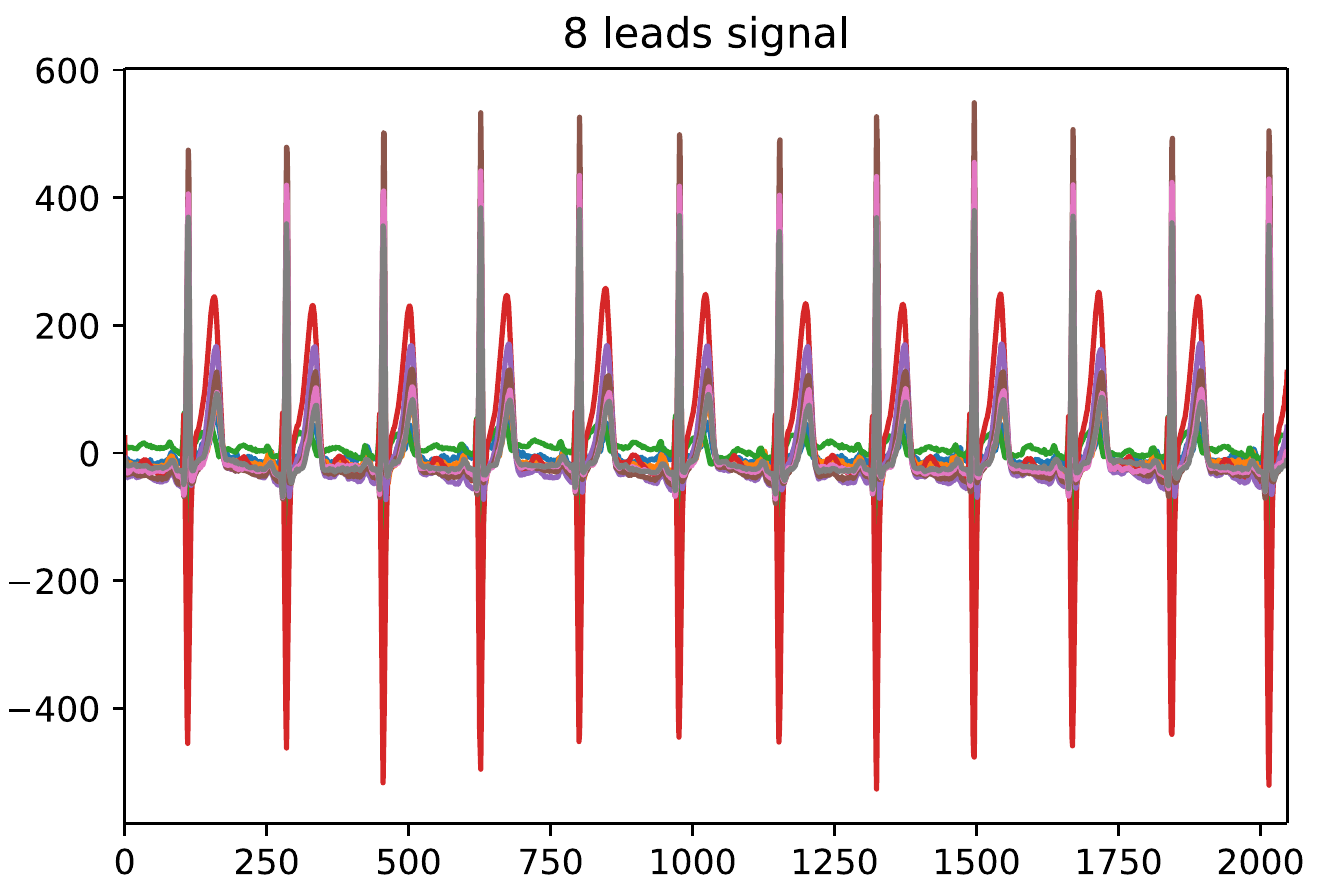}
    \caption{\textbf{The 8-leads} Signal Example from the Alibaba Dataset}
    \label{fig:8lead}
\end{figure}
\begin{figure*}[t] 
\flushleft
\includegraphics[scale=0.72]{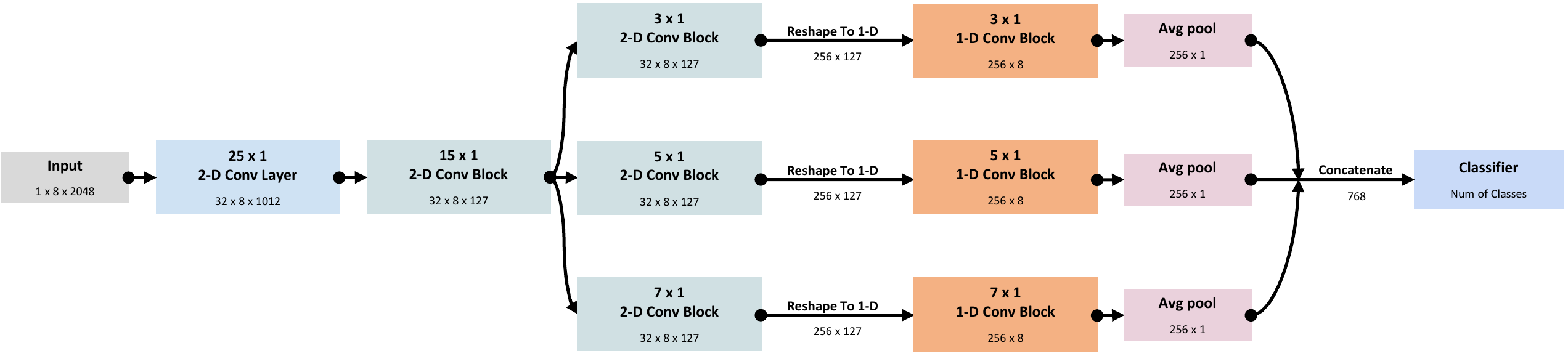}
\caption{\textbf{The Overall Model Structure} }
\label{fig:model}
\end{figure*}

\section{Materials and methods}
\label{sec:model}
\subsection{Dataset}
We adopt two datasets, i.e., the MIT-BIH arrhythmia dataset~\cite{moody2001impact} and the Alibaba arrhythmia dataset, to comprehensively evaluate our proposed model. 

\subsubsection{The MIT-BIH dataset}
\label{MIT-BIH}
The MIT-BIH arrhythmia dataset~\cite{moody2001impact} is a public dataset that has been widely adopted to evaluate the performance of different ECG-based classification algorithms. The MIT-BIH arrhythmia dataset contains record which is 30 minutes, two-leads ECG signal, filtered using a 0.1 to 100HZ bandpass filter. 

Following the standard, mainly because there are only limited amout of arrhythmia cases except these five arrhymia, in this work, we extract 19,481 labelled ECG signal segment containing two leads of five typical kinds of ECG signals, i.e., i.e., N (Normal beat), R (Right bundle branch block beat), A (Atrial premature beat), V (Premature ventricular contraction), L (Left bundle branch block beat).  
The dataset is partitioned into 8:2 for training and testing purposes based on different people.

\subsubsection{The Alibaba dataset}
\label{Alibaba}
The second arrhythmia dataset is provided in the Alibaba ECG signal classification contest. It contains around 20,000 records of 34 different arrhythmia disease. For each record, there are 8 different leads,  i.e.  I, II, V1, V2, V3, V4, V5, V6, sampled at 500 Hz for 10 seconds (5,000 sample points) containing potentially more than one disease. 
One example signal (after re-sampling) is shown in Figure~\ref{fig:8lead}. 
The dataset is partitioned into 8:2 for training and testing purposes based on different people.



\subsection{Problem Definition}
The ECG signal classification task can be formulated into a multi-sequence-to-sequence~\cite{Andrew2017}
task which takes as input a multi-lead ECG signal, which is given in equation~\ref{eq:pro_def}, where n is the number of leads of the ECG signal and m is the length of a single lead. 
\begin{equation}
X =\left(\begin{array}{ccc}
x_{11} & \dots & x_{1 m} \\
\vdots & \vdots & \vdots \\
x_{n 1} & \cdots & x_{n m}
\end{array}\right)
\label{eq:pro_def}
\end{equation}
The output is a sequence of labels, shown in equation~\ref{eq:out}, such that r$_{i}$  represents the possibility of the $i_{th}$ arrhythmia disease.
\begin{equation}
r= [r_{1},....r_{k}]
\label{eq:out}
\end{equation}

The main difficulty of this task is to accurately extract representative features from the ECG signal. Therefore, our approach proposes effective  new model structures for better feature learning. 

\subsection{Data Pre-processing}

 
We adopt the Fast Fourier Transformation (FFT)~\cite{nussbaumer1981fast}  method to pre-process the ECG signals into fixed-length signals mainly because a general classifier requires the input vectors to be of the same length.


The pre-processing procedure is described as follow: 

For each original signal x, the re-sampled signal y is computed through formula~\ref{eq:fft}, where x$_{k}$ is the k-th sample point of the original signal,  y$_{j}$ is the j-th sample point of the re-sampled signal, n is the length of the original signal and i is the imaginary unit. 

\begin{equation}
y_{j}=\sum_{k=1}^{n} x_{k} * \mathrm{e}^{-\frac{2 k i \pi}{n}}   
\label{eq:fft}
\end{equation}

\subsection{Model} 
Figure~\ref{fig:model} shows an overview of our model structure. Our model first extracts features of multi-lead signals with a 2-D convolution layer with a 25*1 convolution kernel  
followed by a 2-D convolution block (with a 15*1 convolution kernel). 
Then we use a layer of parallel 2-D convolution blocks of different kernel sizes (i.e., scale) in order to extract multi-scale features of different leads. 
For each parallel feature extracted from the parallel 2-D convolution block, we reshape the feature into a 1-dimensional feature and use a 1-D convolution block (with the same convolution kernel size) to further extract the feature. 
An average pooling layer is adopted after the 1-D convolution block.  
Lastly, we concatenate features of different scales and use a full connected layer as the classifier to obtain the output.

It is reported by~\cite{hannun2019cardiologist} that a larger convolution kernel is preferred for the early stage of signal feature extraction to capture different stages of the signal in the same cycle. 
As the feature length decreases, we can reduce the size of the convolution kernel to extract fine-grained information. Therefore, in this model, we adopt three different size 2-D convolution kernels and one parallel 1-D convolution kernel for feature extraction. 



\begin{figure}[t]
    \centering
    \includegraphics[scale=0.5]{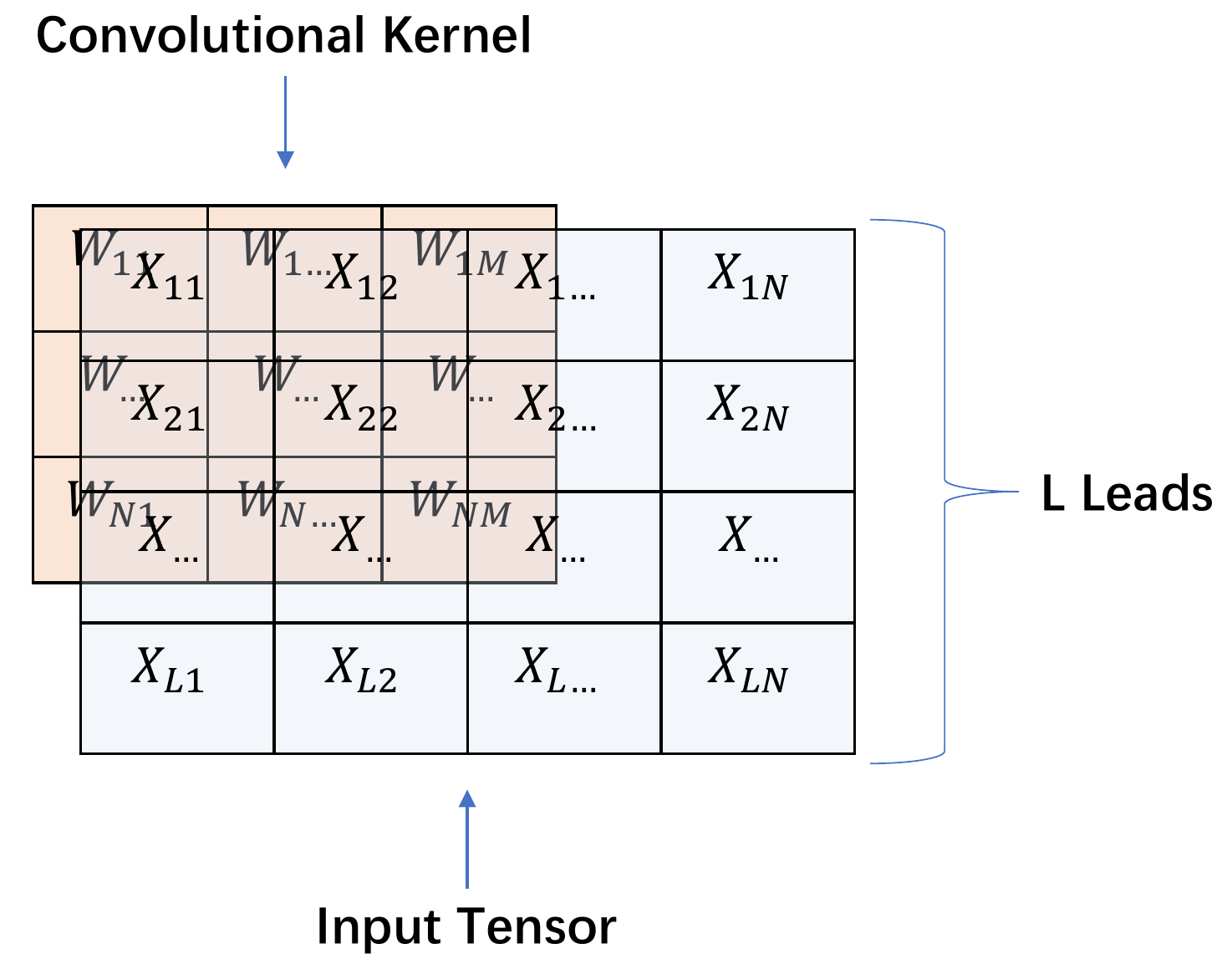}
    \caption{The 2-Dimensional Convolution Operation}
    \label{fig:2doperation}
\end{figure}

\begin{figure}[t]
    \centering
    \includegraphics[scale=0.55]{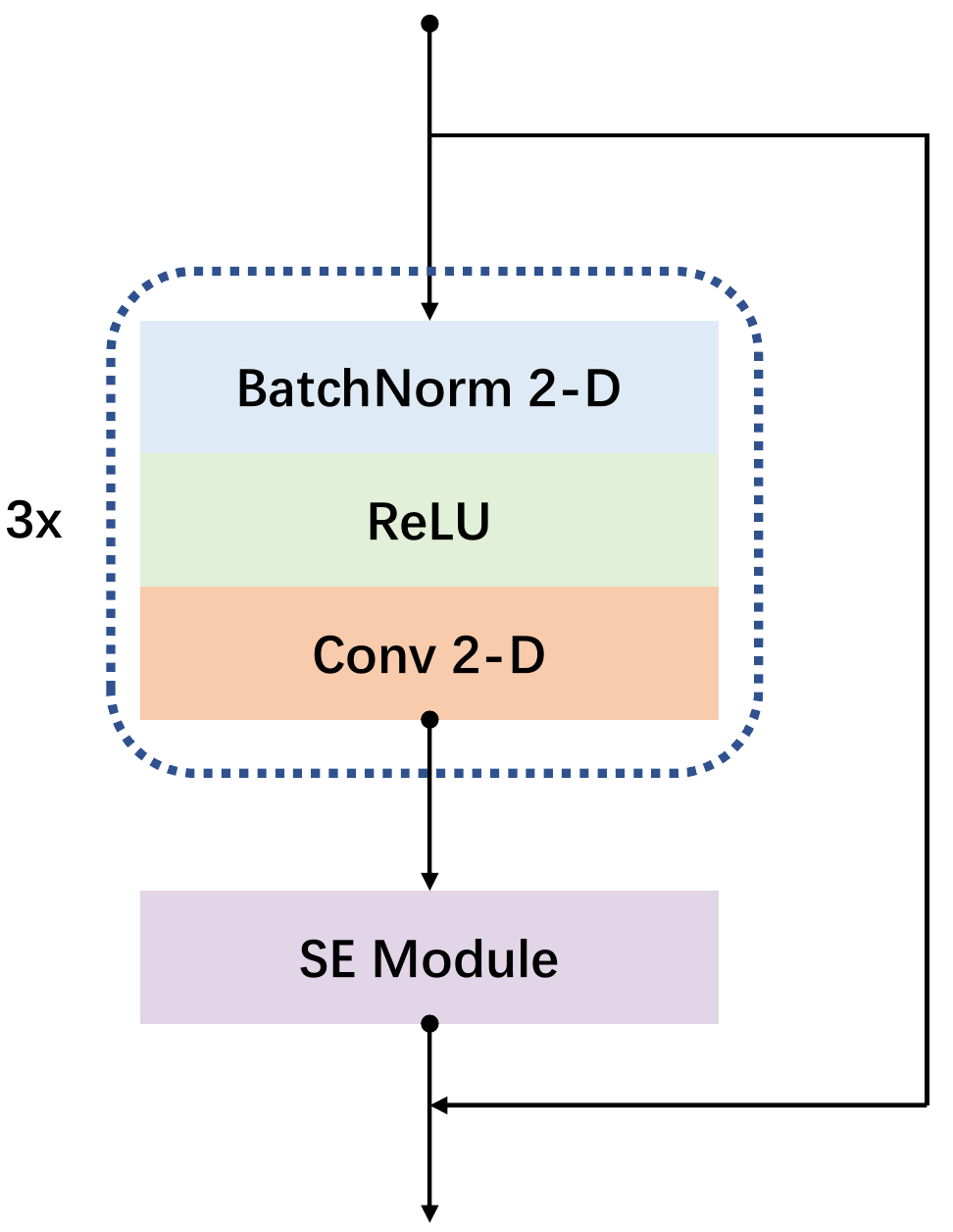}
    \caption{\textbf{ 2-Dimensional CNN Block }}
    \label{fig:2dblock}
\end{figure}

\subsubsection{Layer 1: the 2D-Convolution Layer}
Figure~\ref{fig:8lead} illustrates the ECG signals of 8 leads, distinguished in different colors. Each lead is obtained with a different electrode on a different body part.  
We can observe that the 8 lead signals highly overlap,  which may potentially amplify significant information. 
Therefore, we format the 8-lead signal into a 2-dimensional matrix and adopt the 2-Dimensional convolution layer to learn features from the matrix. 

The 2-Dimensional convolution operation is illustrated in Figure~\ref{fig:2doperation}, which takes the pre-processed signals as a 2-D matrix in order to extract features in different leads. The output of the convolution layer can be represented by Equation.~\ref{eq:operation}:

\begin{equation}
X_{j}^{l }=f\left(\sum_{i \in M_{j}} X_{j}^{l-1} * W_{i j}^{l}+b_{j}^{l}\right)
\label{eq:operation}
\end{equation}

Where $x_{i}^{l}$ is the characteristic vector corresponding to the $j^{t h} \quad$ convolution kernel of the $\quad l^{t h} \quad$ layer and $M_{i} \quad$ is the receptive field of the current neuron, while $W_{i j}^{1}$ indicates the bias coefficient appropriated to the $j^{t h}$ convolution kernel of the $l^{t h}$ layer and $f$ is a nonlinear function.

\subsubsection{Layer 2: the 2-Dimensional Block Layer}
Figure~\ref{fig:2dblock} shows the structure of the 2-D convolution block. 
Following the observations by~\cite{he2016identity}, 
the convolution block in our approach goes through the batch normalization layer and the ReLU layer first before the convolution operation, which has been reported to be a better practice than directly going through the convolution layer. 
	
We adopt the Squeeze-and-Excitation (SE) Module~\cite{hu2018squeeze} to give different weights to different channels of the feature maps, such that the model can learn to focus more on the important features and pay less attention to those less important features. the SE module provides an adaptive way to aggregate features extracted from different leads of the ECG signals, which results in better feature representation. 





\subsubsection{Layer 3: the Parallel 2-D Convolution Blocks Layer} 
It is shown~\cite{szegedy2015going} that different convolution kernel size is able to learn different characteristic representations. 
Combining features extracted with different convolution kernel in parallel provide a better representation of features than using only one convolution kernel. Inspired by this finding, we adopt three parallel paths of convolution blocks of different kernel sizes i.e., 3x1, 5x1, and 7x1, to learn features. 

\begin{figure}[t]
    \centering
    \includegraphics[scale=0.50]{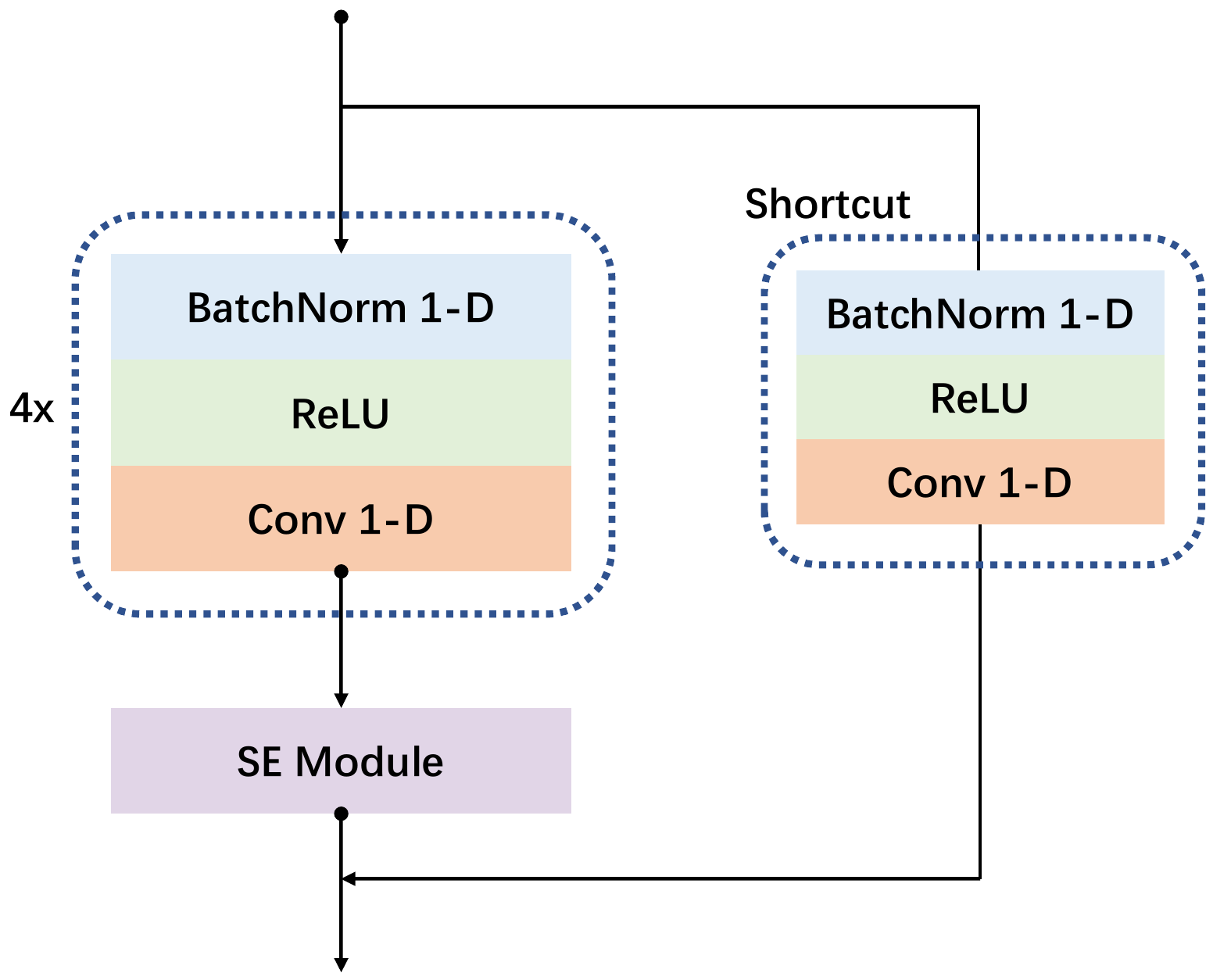}
    \caption{\textbf{ 1-Dimensional CNN Block }}
    \label{fig:1dblock}
\end{figure}

\subsubsection{Layer 4: the Parallel 1-D Convolution Blocks Layer} 
After the 2-D convolution block feature extraction, we begin to focus on more fine-grained features within each signal lead. 
We reshape the features extracted with the 2-D convolution blocks into a 1-Dimensional vector and then further learn features from the 1-D vector using the 1-Dimensional convolution blocks. 

The structure of the 1-D convolution block is shown in Figure~\ref{fig:1dblock}. Similar to the 1-D convolution block, we use a batch normalization layer and a ReLU layer first, followed by a 1-D convolution layer. We also adopt the SE module to aggregate the feature of different channels.   
The 1-D convolution operation is similar to the 2-D convolution operation. It uses a 1-dimensional convolution kernel on a 1-dimensional vector. 

\subsubsection{Layer 5: the Average pooling Layer}
We adopt the standard average pooling layer to downsample the features obtained from each 1-D convolution block, with the purpose of reducing the time in the classifying step and better generalizability of the model.

\subsubsection{Layer 6: the Concatenate and Classify Layer}
Lastly, we concatenate all the features extracted from different parallel convolution blocks and use a fully connected layer as the classifier.




\subsection{The Training Method}


Since the MIT-BIH is a 5-class classification task and the Alibaba dataset is a 34-class classification task, we use different loss functions for them, respectively. 

For the MIT-BIH dataset, we use the standard Cross-Entropy loss function (defined in Equation~\ref{eq:ce}),since the Alibaba dataset has extremely unbalanced data distributions on different categories, we use the Focal Loss function~\cite{lin2017focal} (defined in Equation~\ref{eq:fl}), which have a better performance for the unbalanced categories problem. 
It's known that machine learning approaches usually have trouble learning when one class dominates others. While focal loss can be seen as a kind of hard example mining approach, which reduces the upgraded gradient for those easy examples. The focal loss will make the upgraded gradient of easy example relatively smaller and the one of hard example relatively larger, which is suitable for training networks on unbalanced categories.

\begin{equation}
  \operatorname{Loss}(x, \text { class })=-\log \left(\frac{e^{x[\text {class}]}}{\sum_{j} e^{x[j]}}\right)  
  \label{eq:ce}
\end{equation}
 
 \begin{equation}
 \small
  \operatorname{Loss}(x, \text { class })=-\alpha_{\text {class}}\left(1-\frac{e^{x[\text {class}]}}{\sum_{j} e^{x[j]}}\right)^{\gamma} \log \left(\frac{e^{x[\text {class}]}}{\sum_{j} e^{x[j]}}\right)
  \label{eq:fl}
 \end{equation}

Where $x$ is the output of the model, $class$ is the ground truth label, ${\alpha}_{class}$ is a weighting factor, ${\alpha}$ and ${\gamma}$ are hyper parameters obtained through experiments ${\alpha}= 0.25$, ${\gamma} =2$.


\section{EXPERIMENTS}
\label{sec:exp}
\subsection{Experimental Setup}

\begin{table*}[t]
\centering
\caption{Evaluation Results}
\label{table:1}
\begin{tabular}{c|c|c|c|c|c|c}
{} & \multicolumn {3}{c|}{MIT-BIH dataset} &\multicolumn {3}{c}{Alibaba dataset}\\
{Model} & Precision & Recall & F1-score  & Precision & Recall & F1-score \\ 
Seq2Seq           & 87.7\%    & 87.7\% & 87.7\%         &  33.8\%       & 33.8\%    & 33.8\%       \\ 
ResNet1D-34        & 97.4\%    & 97.4\% & 97.4\%      & 85.7\%    & 85.7\% & 85.7\%   \\ 
ECGNet   & 97.2\%    & 97.2\% & 97.2\%      & 86.2\%    & 86.2\% & 86.2\%   \\ 
\textbf{SE-ECGNet}             & \textbf{99.2\%}    & \textbf{99.2\%} & \textbf{99.2\%}      & \textbf{89.8\%}    & \textbf{89.8\%} & \textbf{89.8\%}   \\ 
\end{tabular}
\end{table*}

We evaluate our model using both the MIT-BIH dataset~\cite{moody2001impact} and Alibaba Contest dataset. 
For both datasets, the FFT method is adopted to pre-process the input signal. 
For the MIT-BIH dataset, we follow the standard and use the five most frequent categories, i.e., N (Normal beat), R (Right bundle branch block beat), A (Atrial premature beat), V (Premature ventricular contraction), L (Left bundle branch block beat), in our evaluation.  For the Alibaba Contest dataset, we use all the 34 classes in our evaluation. 
For both datasets, we partition the training and testing dataset into 4:1 randomly following the standard. 

The maximum epoch is set to 128. The initial learning rate is set to 1e-2, with a learning rate decay strategy. The decay interval of the learning rate is 16, i.e., the decay happens at the 16, 32, 64, 128th epoch, respectively, and we reduce the learning rate by a factor of 10 in each decay interval. With the learning rate decay strategy, we can make the results of network training more stable and closer to its best performance. We use Adam optimizer~\cite{Adam} for weights upgrading. The batch size is set to 256 for the MIT-BIH dataset and 64 for the Alibaba dataset. 


We use the precision, recall, and F1-score as metrics in our evaluation. Especially, in consideration of the complexity of the datasets, we use 5-fold cross-validation on Alibaba dataset. 

We conduct a comprehensive evaluation and compare our approach with 3  state-of-the-art ECG signal classification models, i.e.,  ECGNet~\cite{murugesan2018ecgnet}, Seq2Seq~\cite{mousavi2019inter} and  Resnet-1d~\cite{brito2019electrocardiogram} on both datasets. We adopt the same data pre-processing method and configuration for all methods under the same traning and testing dataset.

We develop all models with PyTorch and run all experiments on an Nvidia GTX 2080Ti 11GB GPU machine.


\subsection{Results and Discussion}
The comparison results are shown in Table~\ref{table:1}. We report the precision, recall, and F1-score on four models evaluated with the two datasets. We highlight the best result for each metric in bold. 
We can observe from the results that our model SE-ECGNet achieves the best performance on all metrics for both datasets. SE-ECGNet achieves 99.2\% on F1-score for the MIT-BIH dataset, which outperforms the other models by around 2\%. For the Alibaba dataset, SE-ECGNet achieves 89.8\% on F1-score, which outperforms the other methods for more than 3\%. 
The Seq2seq model shows the worst performance, especially on the Alibaba dataset, in which only 33.8\% is reported. 
This is mainly due to the information lost during the forward pass of RNN Cells. 
The signal in the Alibaba dataset has a length of more than several thousand data points, and Seq2Seq (which uses LSTM as the backbone) is unable to capture such long dependence information. 
Although  LSTM has solved the problem of long-term dependencies to a certain extent by adding the gating mechanism, it still suffers from poor performance for the extremely long-term dependencies like the signal from the Alibaba dataset. In other words, during the forward pass of LSTM, as the signal is passed forward along the LSTM Cells, some information will be randomly lost regardless of the importance of the information, which results in the poor performance of the Seq2Seq2 model. 

Among the remaining three models, ResNet1D-34 and ECGNet have similar performance in terms of all evaluation metrics on both datasets, respectively. ResNet1D-34 performs 0.2\% better on F1-score on the MIT-BIH dataset and ECGNet performs 0.5\% better on F1-score on the Alibaba dataset. 
One reason for the different performances may due to the different architecture of the two models. 
The architecture of ECGNet is similar to that of the ResNet1D-34, and the difference is that the ECGNet uses LSTM to extract features from the original input data individually and concatenate the features from LSTM with the output of residual blocks. 
Therefore, ECGNet takes the spatial features from CNN and the temporal features from RNN at the same time, and thus ECGNet is able to better capture the feature from the input data. 

Our proposed model SE-ECGNet achieves the best score among all the state-of-the-art models on both datasets.  
The following four advances in our model potentially contribute to performance improvement:  
(1) We adopt the SE\_Res block, which is composed of the residual network and the squeeze-and-excitation Module, which is able to concentrate on the important features; (2) we process the multi-lead input signal into a 2-dimensional matrix and fully utilize the information in the multi-leads; (3) we use multiple convolution layers, with decreasing kernel size, to learn features of different granularity; and (4) we propose the parallel convolution blocks with different kernel size on the same input signal to learn rich features of different scales. 

\begin{table}[t]
\centering
\caption{Ablation Study of Proposed Model}
\label{table:2}
\begin{tabular}{c|c}
Model                    & F1-score        \\ 
\textbf{SE-ECGNet} & \textbf{89.8\%} \\ 
SE-ECGNet w/o 2-D Convolution Blocks            & 86.6\%          \\ 
SE-ECGNet w/o the SE-Module            & 87.4\%          \\ 
SE-ECGNet w/o the Parallel Blocks          & 88.7\%          \\ 
ECGNet           & 86.2\%          \\ 
\end{tabular}
\end{table}

\subsection{The Ablation Study}
To further measure the influences of the proposed advances in our model, we conduct the ablation experiments to explore the effect each individual advance has on the experiment results. We conduct the ablation experiment with the Alibaba dataset since it is more difficult and the performance differences can be clearly observed. 
The ablation experiment results are shown in~\ref{table:2}. We compare the classification results on four different settings, i.e., the SE-ECG full model, the SE-ECGNet without 2-D Convolution Blocks, the SE-ECGNet without the SE-Module and the SE-ECGNet without the Parallel Blocks (with convolution kernel size 5).  

From the experiment results, we can observe that all of the three advances individually improve the performance of the classification. In particular, the 2-D convolution block provides the most improvement, which is 3.2\% on the F1-score. The adoption of the SE-module and the parallel blocks provide 2.4\% and 1.1\% improvement on the F1-score, respectively. Moreover, we can observe that the worst F1-score obtained in the ablation experiment setting, i.e., the SE-ECGNet without 2-D Convolution Blocks, is 0.4\% higher than that of the ECGNet~\cite{murugesan2018ecgnet}, which further shows the improvement of our proposed model structures. 

The greatest improvement that the 2-D convolution block provides on the evaluation results shows that the multi-lead information indeed amplifies the significant features and utilizing the 2-D convolution block is able to effectively learn the features. We can also observe that the SE-module is able to learn high weights on important features, and thus more accurately learn the features. The parallel blocks aggregate the features learned with different convolution kernel size and are able to enhance the final representations learned.

\section{CONCLUSIONS}
\label{sec:conclu}
In this work, we propose a multi-scale deep residual neural network with multi-lead feature extraction, i.e., SE-ECGNet, to classify arrhythmia disease from the ECG signals. 
We are the first to propose to treat the multi-lead signal as a 2-dimensional matrix and combines multi-scale 2-D convolution blocks with 1-D convolution blocks for feature extraction.  We also adopt the squeeze-and-excitation (SE) module to learn proper weights on different channels. 
Our proposed model achieves 99.2\% F1-score in the MIT-BIH dataset and 89.4\% F1-score in Alibaba dataset and outperforms the state-of-the-art performance by 2\% and 3\%, respectively.  \\

\section*{Acknowledgment}

This work has been supported by the National Science Foundation U1836214, 61802275, and Innovation fund of Tianjin Univeristy 2020XRG-0022 

\renewcommand\refname{Reference}

\end{document}